\documentclass[11pt]{article}

\usepackage{acl}
\usepackage{times}
\usepackage{latexsym}
\usepackage[T1]{fontenc}
\usepackage[utf8]{inputenc}
\usepackage{amsmath}
\usepackage{microtype}
\usepackage{inconsolata}
\usepackage{graphicx}
\usepackage{dblfloatfix}
\usepackage{booktabs}
\usepackage{multirow}
\usepackage{enumitem}
\usepackage{xcolor}

\usepackage{tikz}
\usetikzlibrary{arrows.meta, positioning, shapes.geometric}
\usepackage{pgfplots}
\usepgfplotslibrary{polar}
\pgfplotsset{compat=1.18}
\pgfplotsset{compat=1.18}
\usepackage{booktabs}
\usepackage{pgfplots}
\usepackage{cuted}

\title{Case-Aware LLM-as-a-Judge Evaluation for Enterprise-Scale RAG Systems}

\author{
Mukul Chhabra \\
Dell Technologies \\
Austin, TX, USA \\
\texttt{mukul.chhabra@dell.com}
\And
Luigi Medrano \\
Dell Technologies \\
Austin, TX, USA \\
\texttt{Luigi.Medrano@dell.com}
\And
Arush Verma \\
Dell Technologies \\
Austin, TX, USA \\
\texttt{Arush.Verma@dell.com}
}

\begin{document}

\maketitle

\begin{strip}
\begin{abstract}
Enterprise Retrieval-Augmented Generation (RAG) assistants operate in multi-turn, case-based workflows such as technical support and IT operations, where evaluation must reflect operational constraints, structured identifiers (e.g., error codes, versions), and resolution workflows. Existing RAG evaluation frameworks are primarily designed for benchmark-style or single-turn settings and often fail to capture enterprise-specific failure modes such as case misidentification, workflow misalignment, and partial resolution across turns.

We present a case-aware LLM-as-a-Judge evaluation framework for enterprise multi-turn RAG systems. The framework evaluates each turn using eight operationally grounded metrics that separate retrieval quality, grounding fidelity, answer utility, precision integrity, and case/workflow alignment. A severity-aware scoring protocol reduces score inflation and improves diagnostic clarity across heterogeneous enterprise cases. The system uses deterministic prompting with strict JSON outputs, enabling scalable batch evaluation, regression testing, and production monitoring.

Through a comparative study of two instruction-tuned models across short and long workflows, we show that generic proxy metrics provide ambiguous signals, while the proposed framework exposes enterprise-critical tradeoffs that are actionable for system improvement.
\end{abstract}
\end{strip}


\section{Introduction}

Retrieval-Augmented Generation (RAG) is widely used to deploy large language models (LLMs) in enterprise environments by combining retrieval over proprietary content with conditional generation. While effective in reducing hallucinations, enterprise deployments differ substantially from benchmark-style QA: support cases are multi-turn, operationally constrained, and require precise handling of structured identifiers (e.g., error codes, versions) and workflow alignment.

In production systems, responses that appear relevant may still fail to resolve a case, misinterpret structured signals, or violate troubleshooting order. Generic RAG evaluation metrics—such as faithfulness and relevance—often conflate retrieval accuracy, grounding, and resolution quality into coarse signals \citep{es2023ragas,es2024ragasdemo}. As a result, they provide limited diagnostic value for enterprise iteration and deployment decisions.

We propose an enterprise-focused evaluation framework based on the LLM-as-a-Judge paradigm \citep{zheng2023mtbench}. Our key contribution is \textbf{case-aware evaluation}: the judge conditions on multi-turn history, case metadata, and retrieved evidence while enforcing structured scoring across eight enterprise-aligned metrics. The resulting framework exposes operational failure modes—such as retrieval mismatch, hallucination, case misidentification, and workflow misalignment—and provides actionable signals for production monitoring and system improvement.

\subsection{Contributions}

This paper makes the following contributions:

\begin{itemize}[leftmargin=*]
\item We formalize evaluation requirements for enterprise multi-turn RAG systems and identify recurring operational failure modes not captured by generic metric suites.
\item We introduce eight case-aware evaluation metrics that disentangle retrieval quality, grounding fidelity, answer utility, precision integrity, and resolution alignment.
\item We propose a severity-aware scoring protocol that improves diagnostic clarity across heterogeneous enterprise cases.
\item We design an audit-friendly LLM-as-a-Judge implementation with deterministic prompting and structured JSON outputs for scalable batch evaluation.
\item We empirically demonstrate that the proposed framework yields more actionable diagnostic insights than generic proxy metrics.
\end{itemize}

\section{Background and Enterprise Evaluation Requirements}

Enterprise RAG deployments differ from benchmark-style QA in several key respects: queries are multi-turn, retrieval may be partially correct, and answers must comply with operational workflows and structured identifiers (e.g., error codes, versions). In such settings, a response can be relevant yet misaligned with case resolution, or factually grounded yet operationally incorrect.

Generic RAG evaluation frameworks, such as RAGAS \citep{es2023ragas,es2024ragasdemo}, decompose quality into dimensions such as faithfulness and relevance. While effective for reference-free evaluation, these metrics do not explicitly capture enterprise-specific failure modes such as workflow compliance, precision integrity, and correct case interpretation. Our framework complements this line of work by introducing case-aware, operationally grounded evaluation signals.

\subsection{LLM-as-a-Judge for Multi-Turn Evaluation}

LLM-as-a-Judge methods provide scalable evaluation without exhaustive human labeling \citep{zheng2023mtbench}. Prior work identifies potential biases (e.g., verbosity and position bias) and proposes structured rubric-based prompting to improve reliability \citep{liu2023geval,dubois2024lengthcontrolled}. We adopt this paradigm while constraining the judge to evidence-only inputs and enforcing structured JSON outputs to maximize auditability and actionability in enterprise settings.

\section{Related Work}

Automated RAG evaluation has progressed toward decomposed and component-aware metrics. RAGAS introduces reference-free faithfulness and relevance metrics \citep{es2023ragas,es2024ragasdemo}. ARES trains lightweight judges for retrieval and generation scoring \citep{saadfalcon2024ares}, and RAGChecker provides diagnostic benchmarking across retrieval-generation interactions \citep{ru2024ragchecker}. 

Complementary work studies grounding and attribution. RARR proposes retrieval-driven revision pipelines \citep{gao2023rarr}, AttributionBench evaluates attribution reliability \citep{li2024attributionbench}, and FActScore decomposes long-form generations into atomic factual units \citep{min2023factscore}. 

Our framework builds on these directions but targets enterprise RAG systems, introducing case-aware, workflow-sensitive, and precision-focused metrics designed for multi-turn operational support environments.

\section{Framework Overview}

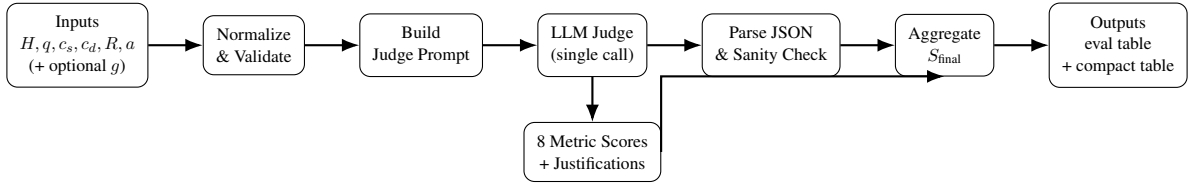
\begin{figure*}[]
\centering
\begin{tikzpicture}[
  scale=0.6, transform shape,
  node distance=8mm and 12mm,
  box/.style={
    draw,
    rounded corners,
    align=center,
    inner sep=8pt,
    font=\normalsize
  },
  arrow/.style={-Latex, thick}
]
\node[box] (inputs) {Inputs\\$H,q,c_s,c_d,R,a$\\(+ optional $g$)};
\node[box, right=of inputs] (norm) {Normalize\\\& Validate};
\node[box, right=of norm] (prompt) {Build\\Judge Prompt};
\node[box, right=of prompt] (judge) {LLM Judge\\(single call)};
\node[box, right=of judge] (parse) {Parse JSON\\\& Sanity Check};
\node[box, below=of judge, yshift=-2mm] (metrics) {8 Metric Scores\\+ Justifications};
\node[box, right=of parse] (agg) {Aggregate\\$S_{\text{final}}$};
\node[box, right=of agg] (out) {Outputs\\eval table\\+ compact table};

\draw[arrow] (inputs) -- (norm);
\draw[arrow] (norm) -- (prompt);
\draw[arrow] (prompt) -- (judge);
\draw[arrow] (judge) -- (parse);
\draw[arrow] (judge) -- (metrics);
\draw[arrow] (parse) -- (agg);
\draw[arrow] (metrics.east) |- (agg.south);
\draw[arrow] (agg) -- (out);
\end{tikzpicture}
\caption{End-to-end evaluation pipeline for case-aware LLM-as-a-Judge scoring of multi-turn RAG responses.}
\label{fig:pipeline}
\end{figure*}

Figure~\ref{fig:pipeline} summarizes the batch evaluation pipeline. Each row corresponds to one evaluated turn with structured case fields, retrieved contexts, and the model answer. For each row, the system normalizes inputs, constructs a single deterministic judge prompt, invokes the LLM once, validates a strict JSON schema, and aggregates per-metric scores into $S_{\text{final}}$.

\paragraph{Inputs and scope.}
Let $q$ be the current query, $H$ the conversation history, $c_s$ the case subject, $c_d$ the case description, $R=\{r_1,\dots,r_n\}$ the retrieved chunks, and $a$ the model answer (optionally a reference $g$). The judge is explicitly constrained to $\{H,q,c_s,c_d,R,a,g\}$ and must not use external knowledge.

\paragraph{Outputs.}
The judge returns eight continuous scores $s_i\in[0,1]$ with brief justifications, plus a weighted aggregate $S_{\text{final}}$ for monitoring and regression tests (0=failure, 1=full compliance; intermediate values reflect partial rubric satisfaction).

\paragraph{Case-aware evaluation.}
We call the evaluation \emph{case-aware} because scoring conditions on (i) case metadata $(c_s,c_d)$, (ii) multi-turn state $H$ capturing prior attempts and constraints, and (iii) retrieved evidence $R$, while restricting judgment to these inputs. This contrasts with common single-turn RAG evaluation that ignores workflow constraints and identifier-critical correctness.

\paragraph{Case-aware evaluation.}
We define \emph{case-aware} LLM-judge evaluation as scoring that conditions on
(i) structured case metadata (e.g., subject/description), (ii) multi-turn history capturing attempted steps and constraints,
and (iii) retrieved evidence, while restricting judgments to these inputs only.
This differs from standard RAG evaluation, which typically assumes single-turn independence and does not model
workflow constraints or identifier-critical correctness.

\subsection{Explainability and Actionability}

Each metric is tied to an engineering lever: \emph{Retrieval Correctness} and \emph{Context Sufficiency} diagnose retriever, chunking, and evidence coverage; \emph{Identifier Integrity} flags high-severity command/ID corruption; and metric-level justifications make regressions auditable and accelerate production triage.

\subsection{Metric Definitions}

We score eight case-aware dimensions spanning evidence quality, grounded response quality, and workflow safety:
\begin{itemize}[leftmargin=*, itemsep=1pt, topsep=2pt]
\item \textbf{Evidence quality:} \emph{Retrieval Correctness} (retrieved chunks contain needed facts) and
\emph{Context Sufficiency} (retrieval covers all required evidence).
\item \textbf{Grounded response quality:} \emph{Hallucination / Grounding Fidelity} (claims supported by $R$),
\emph{Answer Helpfulness} (actionable and clear), and \emph{Answer Type Fit} (diagnose vs.\ instruct vs.\ clarify).
\item \textbf{Workflow safety:} \emph{Identifier Integrity} (no corruption of commands/IDs/paths),
\emph{Case Issue Identification} (correct issue given $H,c_s,c_d$), and \emph{Resolution Alignment} (steps satisfy constraints and likely resolve).
\end{itemize}

Formal definitions and scoring rubrics are provided in Appendix~\ref{app:metrics}.

\section{Severity-Based Scoring and Aggregation}

\subsection{Severity-Based Scoring}
To reflect enterprise risk, composite scores are mapped to four severity bands (Critical, Major, Moderate, Minor); definitions appear in Appendix~\ref{app:severity}.

\subsection{Aggregated Score}
For monitoring and regression testing, we compute a weighted sum:
\begin{equation}
S_{\text{final}} = \sum_{i=1}^{8} w_i s_i,
\label{eq:final_score}
\end{equation}
where weights $w_i$ reflect organizational risk tolerance, emphasizing grounding fidelity and retrieval quality. We use:
Hallucination 0.20, Retrieval Correctness 0.15, Context Sufficiency 0.10, Answer Helpfulness 0.15, Answer Type Fit 0.10, Identifier Integrity 0.10, Case Issue Identification 0.10, Resolution Alignment 0.10.

\section{Implementation}

The framework is a deterministic batch pipeline: each turn is serialized into a structured judge prompt and evaluated with a single LLM call. Outputs must match a strict JSON schema; invalid JSON triggers bounded retries and otherwise fails closed for inspection.

Judge configuration: GPT-4 via Azure OpenAI, temperature=0.0, top\_p=1.0, max\_tokens=1024.

We verify weight-robustness using (i) uniform weights ($w_i{=}0.125$) and (ii) a retrieval-heavy profile (Retrieval Correctness and Context Sufficiency set to 0.20 each; others scaled proportionally). Conclusions are stable: the long-query ordering is unchanged and composite scores shift minimally (typically $<3\%$).


\begin{table*}[t]
\centering
\small
\begin{tabular}{lcccc}
\toprule
\textbf{Metric} & \textbf{GPT-OSS (Short)} & \textbf{LLaMA (Short)} & \textbf{GPT-OSS (Long)} & \textbf{LLaMA (Long)} \\
\midrule
Hallucination (Grounding Fidelity) & 0.6890 & 0.7431 & 0.7586 & 0.7132 \\
Retrieval Correctness & 0.7429 & 0.7647 & 0.7941 & 0.7578 \\
Context Sufficiency & 0.6526 & 0.6429 & 0.7411 & 0.6546 \\
Answer Helpfulness & 0.6943 & 0.6143 & 0.7724 & 0.6143 \\
Answer Type Fit & 0.7603 & 0.7274 & 0.8360 & 0.6825 \\
Identifier Integrity & 0.9421 & 0.9354 & 0.9559 & 0.9329 \\
Case Issue Identification & 0.7636 & 0.6956 & 0.8783 & 0.7052 \\
Resolution Alignment & 0.7004 & 0.6459 & 0.8208 & 0.6759 \\
\midrule
\textbf{Weighted Aggregate} & \textbf{0.7353} & \textbf{0.7202} & \textbf{0.8099} & \textbf{0.7136} \\
\bottomrule
\end{tabular}
\caption{Corrected retrieval-aware evaluation results. Scores reflect full judge evaluation with retrieved contexts provided.}
\label{tab:case_metrics}
\end{table*}

\subsection{Cost Analysis}
Evaluation cost is linear in turns ($O(N)$) with one judge call per turn. Using OpenAI API standard pricing for \texttt{gpt-4.1-2025-04-14} (\$3 / 1M input tokens, \$12 / 1M output tokens) \citep{openai_pricing}, the per-turn judge cost is
$\approx 3\!\times\!10^{-6}T_{\text{in}} + 12\!\times\!10^{-6}T_{\text{out}}$ dollars.
For a typical prompt of $\sim$3{,}000 input tokens and $\sim$400 output tokens, this is $\approx\$0.014$ per turn; evaluating 469 turns would cost $\approx\$6.6$ (Azure pricing may differ, but the estimate is directly token-scalable).
\section{Evaluation Methodology}

This section describes the experimental protocol used to evaluate the proposed case-aware LLM-as-a-Judge framework within an enterprise troubleshooting setting.

\subsection{Dataset and Experimental Setup}

Evaluation was conducted on two disjoint subsets of anonymized enterprise support cases:

\begin{itemize}[leftmargin=*]
    \item \textbf{Short queries:} 237 cases
    \item \textbf{Long queries:} 232 cases
\end{itemize}

Each row corresponds to a single evaluated turn (one model response) with any available prior conversation history included in the case fields. The judge is invoked once per row/turn. When reporting conversation-level results, we aggregate turn-level scores by a conversation identifier.

Cases were sampled from real-world enterprise troubleshooting scenarios and normalized to remove personally identifiable information.
\paragraph{Sampling Procedure.}
Cases were sampled from anonymized enterprise support logs across multiple product categories over a six-month period.
Sampling was stratified to include both short-form queries (single-issue diagnostic requests) and long-form diagnostic
narratives containing prior troubleshooting context. Duplicate cases and non-technical requests were excluded.

\subsection{Evaluation Paradigms}

We compare three evaluation approaches:

\begin{itemize}[leftmargin=*]
\item \textbf{Generic proxy metrics:} RAGAS-style signals including faithfulness and answer relevancy.
\item \textbf{Heuristic checks:} keyword overlap and retrieval scoring thresholds.
\item \textbf{Proposed framework:} an eight-metric case-aware judge with severity-weighted aggregation into a final score $S_{final}$.
\end{itemize}

The objective is to assess whether enterprise-aligned metrics provide more diagnostic and operationally actionable signals than generic proxies.

\subsection{Models Evaluated}

To stress-test the evaluation framework across heterogeneous model behavior, we evaluated two instruction-tuned models representing distinct architectural trade-offs.

All models were tested using identical retrieval pipelines, prompts, and conversation histories to isolate model and evaluation effects from system configuration differences.

\begin{itemize}[leftmargin=*]
\item \textbf{Llama-3.3-70B-Instruct}
\item \textbf{gpt-oss-120b}
\end{itemize}

The comparison is intentionally heterogeneous to demonstrate that the framework remains discriminative across model families.

\section{Results and Analysis}

This evaluation does not seek leaderboard ranking, but rather tests whether case-aware metrics reveal enterprise-relevant distinctions that generic proxy signals overlook.

Table~\ref{tab:case_metrics} reports mean enterprise metric scores across evaluated cases.
\begin{table*}[t]
\centering
\small
\setlength{\tabcolsep}{5pt}
\begin{tabular}{lccccc}
\toprule
\textbf{Dataset} & \textbf{n} & \textbf{Mean (GPT)} & \textbf{Mean (Llama)} & $\boldsymbol{\Delta}$ & \textbf{Wilcoxon $p$} \\
\midrule
Short Queries & 70 & 0.7353 & 0.7202 & +0.0151 & 0.6495 \\
Long Queries  & 63 & 0.8099 & 0.7136 & +0.0963 & 0.0011 \\
\bottomrule
\end{tabular}
\caption{Paired Wilcoxon signed-rank test on per-case weighted aggregate scores. $\Delta$ denotes GPT minus Llama.}
\label{tab:significance}
\end{table*}
\subsection{Key Findings}

Proxy metrics produce mixed signals. In contrast, the proposed case-aware framework exposes workflow-critical differences including issue targeting, resolution alignment, and precision integrity.

\begin{itemize}[leftmargin=*]
\item On short queries, Llama demonstrates stronger grounding fidelity and identifier preservation.
\item GPT-oss exhibits higher helpfulness and stronger alignment with case resolution workflows.
\item On long-form diagnostic queries, GPT-oss achieves significantly higher weighted aggregate scores, indicating greater robustness under context-heavy and multi-step diagnostic conditions.
\end{itemize}

These findings demonstrate that faithfulness and relevance alone are insufficient for enterprise deployment decisions.

\subsection{Input Complexity}

Short queries average 3.47 words (median = 3), whereas long queries average 14.76 words (median = 14), representing a 4.25$\times$ increase in input length. The long-query subset therefore reflects substantially higher descriptive complexity and retrieval demands.

\subsection{Statistical Significance}
\paragraph{Unit of analysis.}
Tests are performed at the \emph{conversation} (case) level, not per turn. We group turns by conversation identifier and
compute each conversation’s mean weighted composite score. This yields $n{=}70$ short-query conversations and $n{=}63$
long-query conversations, avoiding inflated significance from within-conversation correlation.

\paragraph{Tests.}
We use paired, two-sided Wilcoxon signed-rank tests on per-conversation weighted scores (Table~\ref{tab:significance});
paired $t$-tests and bootstrap confidence intervals are reported as robustness checks.

For short queries, GPT-oss achieves a directionally higher weighted aggregate score, but the difference is not statistically significant ($p=0.6495$). The bootstrap 95\% confidence interval for the mean difference is $[-0.0315, 0.0623]$, indicating overlap around zero.

For long diagnostic queries, GPT-oss significantly outperforms Llama under the case-aware evaluation framework ($p=0.0011$). The bootstrap 95\% confidence interval for the mean difference is $[0.0486, 0.1469]$, remaining strictly above zero. This indicates that performance separation becomes pronounced under higher input complexity and multi-step diagnostic reasoning.

\subsection{Judge Robustness}
We re-ran the full case-aware evaluation with \texttt{llama-3.3-70b-instruct} as an alternative judge using the same
rubric, strict JSON schema, and aggregation weights (Eq.~\ref{eq:final_score}). This second judge is not treated as
ground truth; it is used to probe stability under evaluator variation.

Conclusions are stable at the system level: GPT-oss scores higher than Llama under both judges (Table~\ref{tab:judge_robustness}).
The long-query separation remains statistically significant (GPT-4 judge: $p{=}0.0011$; Llama judge: $p{=}0.0016$).
Short-query effects are smaller and judge-sensitive (GPT-4: $p{=}0.6495$; Llama: $p{=}0.0005$). Inter-judge agreement on the
weighted aggregate is moderate-to-strong ($\rho \in [0.52, 0.69]$), with lower agreement on a few short-query metrics
(Appendix~\ref{app:judge_robustness}).

\begin{table*}[t]
\centering
\small
\setlength{\tabcolsep}{4pt}
\begin{tabular}{llccccc}
\toprule
Subset & Judge & $n$ & Mean (GPT-oss) & Mean (Llama) & $\Delta$ & Wilcoxon $p$ \\
\midrule
Short & GPT-4 & 70 & 0.7353 & 0.7202 & +0.0151 & 0.6495 \\
Short & Llama-3.3-70B & 70 & 0.6639 & 0.6060 & +0.0578 & 0.0005 \\
Long & GPT-4 & 63 & 0.8099 & 0.7136 & +0.0963 & 0.0011 \\
Long & Llama-3.3-70B & 63 & 0.7143 & 0.6589 & +0.0554 & 0.0016 \\
\bottomrule
\end{tabular}
\caption{Judge-robustness check on conversation-level weighted aggregate scores ($S_{\text{final}}$). $\Delta$ denotes GPT-oss minus Llama.}
\label{tab:judge_robustness}
\end{table*}

\subsection{Comparison with Generic Proxy Metrics}

Proxy metrics suggested higher faithfulness for Llama and stronger answer relevancy for GPT-oss (Table~\ref{tab:proxy}), yet failed to indicate which system better supported case resolution workflows.

\begin{table}[t]
\centering
\small
\begin{tabular}{lcc}
\toprule
Metric & Llama & GPT-oss \\
\midrule
Faithfulness (Short) & 0.8439 & 0.5799 \\
Answer Relevancy (Short) & 0.7404 & 0.8192 \\
Faithfulness (Long) & 0.8123 & 0.6005 \\
Answer Relevancy (Long) & 0.7276 & 0.8330 \\
\bottomrule
\end{tabular}
\caption{Representative proxy metrics produce mixed signals and do not capture workflow alignment or issue targeting.}
\label{tab:proxy}
\end{table}

Unlike proxy metrics, the proposed framework decomposes performance into actionable enterprise dimensions, reducing the risk of false confidence during deployment decisions.

\subsection{Enterprise Diagnosis and Operational Use}

Metric-level outputs enable targeted engineering intervention:

\begin{itemize}[leftmargin=*]
\item Low retrieval correctness with low sufficiency indicates retriever or chunk refinement weaknesses.
\item Strong grounding but weak helpfulness suggests response structuring improvements.
\item Low alignment signals workflow violations requiring improved case extraction or conversational memory.
\end{itemize}

Severity-weighted scoring prevents inflation from partially correct answers, ensuring high-risk failures meaningfully impact aggregate evaluation.

This enables:

\begin{itemize}[leftmargin=*]
\item Release gating via thresholded $S_{final}$
\item Regression testing across model updates
\item Root-cause diagnosis through metric decomposition
\item Continuous production monitoring
\end{itemize}

\subsection{Human Alignment Validation}

To assess alignment between the LLM judge and human evaluators, we conducted a lightweight validation on a
stratified random sample of 60 evaluated turns (30 short, 30 long). Two domain experts independently reviewed
each sampled turn using the same case fields and retrieved contexts available to the judge.

Humans provided binary judgments for three operationally critical dimensions: Hallucination, Identifier Integrity,
and Resolution Alignment. Agreement between the LLM judge and the majority human vote was:

\begin{itemize}[leftmargin=*]
    \item Hallucination: 88\%
    \item Identifier Integrity: 91\%
    \item Resolution Alignment: 84\%
\end{itemize}

Disagreements occurred primarily in borderline cases involving partially correct remediation steps or implicitly
referenced identifiers. Overall, results suggest strong alignment for high-risk enterprise failure modes.

\section{Limitations}
Our framework relies on the availability of representative evaluation conversations and high-quality case fields (subject/description) to enable case-aware judging. While severity-based scoring improves stability for heterogeneous enterprise cases, rubric calibration and metric weights may require adjustment across different organizations, industries, and risk tolerances.

\section{Conclusion}
We present a practical, case-aware LLM-as-a-judge evaluation framework for enterprise multi-turn RAG systems. By grounding evaluation in operational failure modes including workflow alignment, issue targeting, and precision integrity, the framework provides diagnostic signals that generic proxy metrics often miss. The resulting metric suite and batch implementation enable scalable auditing, regression testing, and release gating decisions rooted in production realities.

\section{Reproducibility and Artifacts}
To support replication, we document the full evaluation protocol (judge rubric, prompt structure, JSON schema, and
aggregation procedure) and provide code/configuration details sufficient to reproduce the reported aggregates on any
similarly formatted case set. Due to enterprise confidentiality, raw case logs cannot be shared. Where permitted, we
will make available de-identified examples and/or synthetic cases that preserve identifier and workflow structure to
illustrate end-to-end scoring and aggregation.

\noindent\textbf{Available (where permitted):} judge rubric and prompt templates; strict JSON schema and validation rules;
aggregation weights/config profiles; analysis and significance-testing scripts; de-identified and/or synthetic example
cases with expected judge outputs.

\section{Ethical and Safety Considerations}
Enterprise support assistants can recommend operational actions with real-world consequences.
Our evaluation explicitly treats unsafe or destructive guidance as severe failures via hallucination (grounding fidelity), identifier integrity, and resolution alignment. The framework is intended to support release gating and regression
testing to reduce deployment risk. We also note that LLM-judge evaluation can encode biases; we mitigate this via
deterministic prompting, strict evidence-only inputs, schema validation, and (when feasible) cross-judge and human
alignment checks.


\nocite{*}
\bibliography{custom}

@inproceedings{lewis2020rag,
  title     = {Retrieval-Augmented Generation for Knowledge-Intensive {NLP} Tasks},
  author    = {Lewis, Patrick and Perez, Ethan and Piktus, Aleksandra and Petroni, Fabio and Karpukhin, Vladimir and Goyal, Naman and K{\"u}ttler, Heinrich and Lewis, Mike and Yih, Wen{-}tau and Rockt{\"a}schel, Tim and Riedel, Sebastian},
  booktitle = {Advances in Neural Information Processing Systems},
  year      = {2020}
}

@article{es2023ragas,
  title   = {{RAGAS}: Automated Evaluation of Retrieval-Augmented Generation},
  author  = {Es, Shahul and James, Jithin and Espinosa-Anke, Luis and Schockaert, Steven},
  journal = {arXiv preprint arXiv:2309.15217},
  year    = {2023},
  url     = {https://arxiv.org/abs/2309.15217}
}

@inproceedings{es2024ragasdemo,
  title     = {{RAGAs}: Automated Evaluation of Retrieval-Augmented Generation},
  author    = {Es, Shahul and James, Jithin and Espinosa-Anke, Luis and Schockaert, Steven},
  booktitle = {Proceedings of the 18th Conference of the European Chapter of the Association for Computational Linguistics: System Demonstrations},
  year      = {2024}
}

@article{zheng2023mtbench,
  title   = {Judging {LLM}-as-a-Judge with {MT}-Bench and Chatbot Arena},
  author  = {Zheng, Lianmin and Chiang, Wei-Lin and Sheng, Ying and Zhuang, Siyuan and Wu, Zhanghao and Zhuang, Yonghao and Lin, Zi and Li, Zhuohan and Li, Dacheng and Xing, Eric P. and Zhang, Hao and Gonzalez, Joseph E. and Stoica, Ion},
  journal = {arXiv preprint arXiv:2306.05685},
  year    = {2023},
  url     = {https://arxiv.org/abs/2306.05685}
}

@inproceedings{saadfalcon2024ares,
  title     = {{ARES}: An Automated Evaluation Framework for Retrieval-Augmented Generation Systems},
  author    = {Saad-Falcon, Jon and Khattab, Omar and Potts, Christopher and Zaharia, Matei},
  booktitle = {Proceedings of the 2024 Conference of the North American Chapter of the Association for Computational Linguistics: Human Language Technologies (Volume 1: Long Papers)},
  year      = {2024},
  month     = jun,
  address   = {Mexico City, Mexico},
  publisher = {Association for Computational Linguistics},
  pages     = {338--354},
  url       = {https://aclanthology.org/2024.naacl-long.20/},
  doi       = {10.18653/v1/2024.naacl-long.20}
}

@inproceedings{ru2024ragchecker,
  title     = {{RAGChecker}: A Fine-grained Framework for Diagnosing Retrieval-Augmented Generation},
  author    = {Ru, Dongyu and Qiu, Lin and Hu, Xiangkun and Zhang, Tianhang and Shi, Peng and Chang, Shuaichen and Jiayang, Cheng and Wang, Cunxiang and Sun, Shichao and Li, Huanyu and Zhang, Zizhao and Wang, Binjie and Jiang, Jiarong and He, Tong and Wang, Zhiguo and Liu, Pengfei and Zhang, Yue and Zhang, Zheng},
  booktitle = {Advances in Neural Information Processing Systems 37 (NeurIPS 2024), Datasets and Benchmarks Track},
  year      = {2024},
  url       = {https://papers.nips.cc/paper_files/paper/2024/hash/27245589131d17368cccdfa990cbf16e-Abstract-Datasets_and_Benchmarks_Track.html},
  doi       = {10.52202/079017-0692}
}

@inproceedings{gao2023rarr,
  title     = {{RARR}: Researching and Revising What Language Models Say, Using Language Models},
  author    = {Gao, Luyu and Dai, Zhuyun and Pasupat, Panupong and Chen, Anthony and Chaganty, Arun Tejasvi and Fan, Yicheng and Zhao, Vincent and Lao, Ni and Lee, Hongrae and Juan, Da-Cheng and Guu, Kelvin},
  booktitle = {Proceedings of the 61st Annual Meeting of the Association for Computational Linguistics (Volume 1: Long Papers)},
  year      = {2023},
  month     = jul,
  address   = {Toronto, Canada},
  publisher = {Association for Computational Linguistics},
  pages     = {16477--16508},
  url       = {https://aclanthology.org/2023.acl-long.910/},
  doi       = {10.18653/v1/2023.acl-long.910}
}

@inproceedings{li2024attributionbench,
  title     = {{A}ttribution{B}ench: How Hard is Automatic Attribution Evaluation?},
  author    = {Li, Yifei and Yue, Xiang and Liao, Zeyi and Sun, Huan},
  booktitle = {Findings of the Association for Computational Linguistics: ACL 2024},
  year      = {2024},
  month     = aug,
  address   = {Bangkok, Thailand},
  publisher = {Association for Computational Linguistics},
  pages     = {14919--14935},
  url       = {https://aclanthology.org/2024.findings-acl.886/},
  doi       = {10.18653/v1/2024.findings-acl.886}
}

@inproceedings{min2023factscore,
  title     = {{FActScore}: Fine-Grained Atomic Evaluation of Factual Precision in Long-Form Text Generation},
  author    = {Min, Sewon and Krishna, Kalpesh and Lyu, Xinxi and Lewis, Mike and Yih, Wen{-}tau and Koh, Pang Wei and Iyyer, Mohit and Zettlemoyer, Luke and Hajishirzi, Hannaneh},
  booktitle = {Proceedings of the 2023 Conference on Empirical Methods in Natural Language Processing},
  year      = {2023}
}

@article{manakul2023selfcheckgpt,
  title   = {{SelfCheckGPT}: Zero-Resource Black-Box Hallucination Detection for Generative Large Language Models},
  author  = {Manakul, Potsawee and Liusie, Adian and Gales, Mark J. F.},
  journal = {arXiv preprint arXiv:2303.08896},
  year    = {2023},
  url     = {https://arxiv.org/abs/2303.08896}
}

@inproceedings{liu2023geval,
  title     = {{G}-Eval: {NLG} Evaluation using Gpt-4 with Better Human Alignment},
  author    = {Liu, Yang and Iter, Dan and Xu, Yichong and Wang, Shuohang and Xu, Ruochen and Zhu, Chenguang},
  booktitle = {Proceedings of the 2023 Conference on Empirical Methods in Natural Language Processing},
  year      = {2023},
  month     = dec,
  address   = {Singapore},
  publisher = {Association for Computational Linguistics},
  pages     = {2511--2522},
  url       = {https://aclanthology.org/2023.emnlp-main.153/},
  doi       = {10.18653/v1/2023.emnlp-main.153}
}

@article{dubois2024lengthcontrolled,
  title   = {Length-Controlled AlpacaEval: A Simple Way to Debias Automatic Evaluators},
  author  = {Dubois, Yann and Galambosi, Bal{\'a}zs and Liang, Percy and Hashimoto, Tatsunori B.},
  journal = {arXiv preprint arXiv:2404.04475},
  year    = {2024},
  url     = {https://arxiv.org/abs/2404.04475}
}

@article{liang2022helm,
  title   = {Holistic Evaluation of Language Models},
  author  = {Liang, Percy and Bommasani, Rishi and Lee, Tony and Tsipras, Dimitris and Soylu, Dilara and Yasunaga, Michihiro and Zhang, Yian and Narayanan, Deepak and Wu, Yuhuai and Kumar, Ananya and Newman, Benjamin and Yuan, Binhang and Yan, Bobby and Zhang, Ce and Cosgrove, Christian and Manning, Christopher D. and R{\'e}, Christopher and Acosta-Navas, Diana and Hudson, Drew A. and Zelikman, Eric and Durmus, Esin and Ladhak, Faisal and Rong, Frieda and Ren, Hongyu and Yao, Huaxiu and Wang, Jue and Santhanam, Keshav and Orr, Laurel and Zheng, Lucia and Yuksekgonul, Mert and Suzgun, Mirac and Kim, Nathan and Guha, Neel and Chatterji, Niladri and Khattab, Omar and Henderson, Peter and Huang, Qian and Chi, Ryan and Xie, Sang Michael and Santurkar, Shibani and Ganguli, Surya and Hashimoto, Tatsunori and Icard, Thomas and Zhang, Tianyi and Chaudhary, Vishrav and Wang, William and Li, Xuechen and Mai, Yifan and Zhang, Yuhui and Koreeda, Yuta},
  journal = {arXiv preprint arXiv:2211.09110},
  year    = {2022},
  url     = {https://arxiv.org/abs/2211.09110}
}

@inproceedings{lin2022truthfulqa,
  title     = {{TruthfulQA}: Measuring How Models Mimic Human Falsehoods},
  author    = {Lin, Stephanie and Hilton, Jacob and Evans, Owain},
  booktitle = {Proceedings of the 60th Annual Meeting of the Association for Computational Linguistics},
  year      = {2022}
}

@inproceedings{mehri2020usereval,
  title     = {{USR}: An Unsupervised and Reference-Free Evaluation Metric for Dialog Generation},
  author    = {Mehri, Shikib and Eskenazi, Maxine},
  booktitle = {Proceedings of the 58th Annual Meeting of the Association for Computational Linguistics},
  year      = {2020}
}

@inproceedings{ribeiro2020checklist,
  title     = {Beyond Accuracy: Behavioral Testing of {NLP} Models with {C}heck{L}ist},
  author    = {Ribeiro, Marco Tulio and Wu, Tongshuang and Guestrin, Carlos and Singh, Sameer},
  booktitle = {Proceedings of the 58th Annual Meeting of the Association for Computational Linguistics},
  year      = {2020},
  month     = jul,
  address   = {Online},
  publisher = {Association for Computational Linguistics},
  pages     = {4902--4912},
  url       = {https://aclanthology.org/2020.acl-main.442/},
  doi       = {10.18653/v1/2020.acl-main.442}
}

@misc{openai_pricing,
  title        = {OpenAI API Pricing},
  author       = {{OpenAI}},
  year         = {2026},
  howpublished = {\url{https://developers.openai.com/api/docs/pricing/}},
  note         = {Accessed: 2026-02-14}
}

@article{nath2025domain,
  title   = {Domain-Adaptive Small Language Models for Structured Tax Code Prediction},
  author  = {Souvik Nath and Sumit Wadhwa and Luis Perez},
  journal = {arXiv preprint arXiv:2507.10880},
  year    = {2025},
  url     = {https://arxiv.org/abs/2507.10880}
}

@article{nema2025modp,
  title   = {MODP: Multi Objective Directional Prompting},
  author  = {Aashutosh Nema and Samaksh Gulati and Evangelos Giakoumakis and Bipana Thapaliya},
  journal = {arXiv preprint arXiv:2504.18722},
  year    = {2025},
  url     = {https://arxiv.org/abs/2504.18722}
}

\clearpage

\begin{figure*}[!b]
\centering
\begin{tikzpicture}
\begin{axis}[
  ybar,
  bar width=8pt,
  height=5.0cm,
  width=0.95\textwidth,
  ymin=0, ymax=0.25,
  ylabel={Weight},
  symbolic x coords={Halluc.,Retr.,Suff.,Help.,TypeFit,Ident.,IssueID,Align.},
  xtick=data,
  x tick label style={rotate=25, anchor=east, font=\scriptsize},
  y tick label style={font=\scriptsize},
  ylabel style={font=\small},
  title style={font=\small},
  title={Example default aggregation weights (tune per domain/risk)},
  grid=both,
  major grid style={draw=gray!20},
]
\addplot coordinates {
  (Halluc.,0.20)
  (Retr.,0.15)
  (Suff.,0.10)
  (Help.,0.15)
  (TypeFit,0.10)
  (Ident.,0.12)
  (IssueID,0.09)
  (Align.,0.09)
};
\end{axis}
\end{tikzpicture}
\caption{Example metric weights for $S_{\text{final}}$. Rendered as a wide figure to avoid cut-off in two-column layout.}
\label{fig:weights}
\end{figure*}
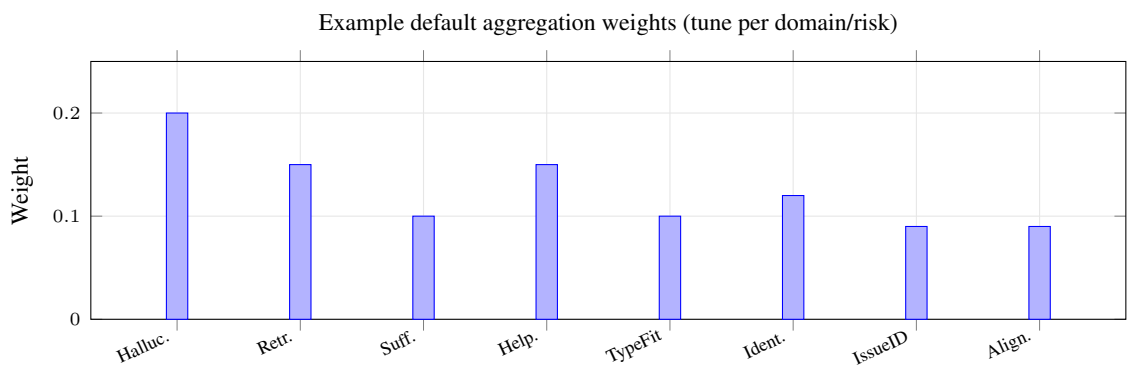
\appendix

\section{Extended Experimental Analysis}
\label{sec:appendix_extended}

This appendix provides a deeper analysis of the proposed case-aware LLM-as-a-judge framework. The main paper presents high-level results and key findings; here we include diagnostic visualizations, statistical observations, and implementation specifics that further validate robustness, interpretability, and enterprise applicability.

\subsection{Evaluation Dataset Characteristics}

The evaluation dataset consists of multi-turn enterprise support conversations involving troubleshooting, configuration issues, and error-code-driven resolution workflows.

Each conversation turn was independently evaluated across eight continuous metrics:

\begin{itemize}
    \item Hallucination
    \item Retrieval Correctness
    \item Context Sufficiency
    \item Answer Helpfulness
    \item Answer Type Fit
    \item Identifier Integrity
    \item Case Issue Identification
    \item Resolution Alignment
\end{itemize}

Each metric is represented as a continuous scalar in $[0,1]$. Scores near 0 indicate severe rubric violations, while values near 1 indicate strong compliance. Intermediate values represent partial satisfaction of the metric criteria.

\subsection{Operational Evaluation Considerations}

Enterprise evaluation systems require strict parsing and validation, including JSON-only outputs, bounded score ranges, and required metric keys.

Batch scoring may require extended runtime; therefore progress tracking, retry mechanisms, and output logging are necessary for reliability. All prompts, raw judge outputs, and parsed metric scores were stored for auditability.

For diagnostic workflows, two output formats are recommended:

\begin{itemize}[leftmargin=*]
\item \textbf{Granular output:} per-row metric scores, textual justifications, and raw judge outputs.
\item \textbf{Compact output:} case fields, response, retrieved contexts, and aggregated $S_{\text{final}}$ for monitoring dashboards.
\end{itemize}

\section{Severity Band Definitions}
\label{app:severity}
\subsection{Severity-Based Scoring}

In enterprise settings, a single severe failure (e.g., hallucinated destructive action) should strongly impact evaluation. We therefore recommend severity-based scoring bands, where the judge first identifies the most severe issue per metric and maps it to a score region:

\begin{itemize}[leftmargin=*]
\item \textbf{Severe} failure: 0.00--0.30 (unsafe, contradictory, or clearly wrong)
\item \textbf{Moderate} failure: 0.31--0.60 (partly correct but flawed / incomplete)
\item \textbf{Minor} issues: 0.61--0.85 (mostly correct with small gaps)
\item \textbf{No meaningful issues}: 0.86--1.00 (excellent)
\end{itemize}

This reduces score inflation, improves stability across heterogeneous cases, and makes metric outputs more interpretable by engineering teams.

\section{Enterprise Metric Suite}
\label{app:metrics}

We define eight \textbf{independent} and \textbf{actionable} metrics designed to disentangle retrieval quality, grounding fidelity, resolution utility, precision integrity, and workflow alignment. Each metric maps directly to a distinct engineering intervention.

\paragraph{M1: Hallucination (Grounding Fidelity).}
Measures whether claims and prescribed steps are supported by retrieved context and case fields. Unsupported or contradictory assertions reduce this score.

\paragraph{M2: Retrieval Correctness.}
Evaluates whether retrieved contexts $R$ are relevant and case-appropriate, isolating retriever-layer failures such as incorrect versions, environments, or stale documentation.

\paragraph{M3: Context Sufficiency.}
Assesses whether $R$ provides enough evidence for a safe response. When evidence is incomplete, strong answers request clarification rather than speculate.

\paragraph{M4: Answer Helpfulness.}
Captures practical utility: whether the response advances resolution through prioritized actions, structured troubleshooting, or clarifying questions.

\paragraph{M5: Answer Type Fit.}
Determines whether response specificity matches query intent (e.g., concrete commands for operational issues versus high-level guidance for conceptual questions).

\paragraph{M6: Identifier Integrity.}
Evaluates correct interpretation and preservation of structured identifiers (e.g., error codes, versions, commands). Corruption or misapplication is penalized.

\paragraph{M7: Case Issue Identification.}
Measures whether the response correctly identifies the underlying issue rather than drifting into adjacent topics.

\paragraph{M8: Resolution Alignment.}
Assesses compliance with case constraints and workflow, including respect for prior attempted steps, required sequencing, and prohibited actions.

\section{Dataset Composition Details}

The evaluation dataset spans multiple enterprise storage and infrastructure product families across heterogeneous troubleshooting workflows.

Error-code-driven cases represent approximately 3.4\% of short cases and 6.0\% of long cases.

All cases were sampled from anonymized enterprise troubleshooting scenarios and normalized to remove personally identifiable information.

\subsection{Inter-Metric Correlation Analysis}

To examine metric redundancy and interdependence, we compute Pearson correlation coefficients across all eight metrics.

\begin{figure}[t]
    \centering
    \includegraphics[width=\linewidth]{ 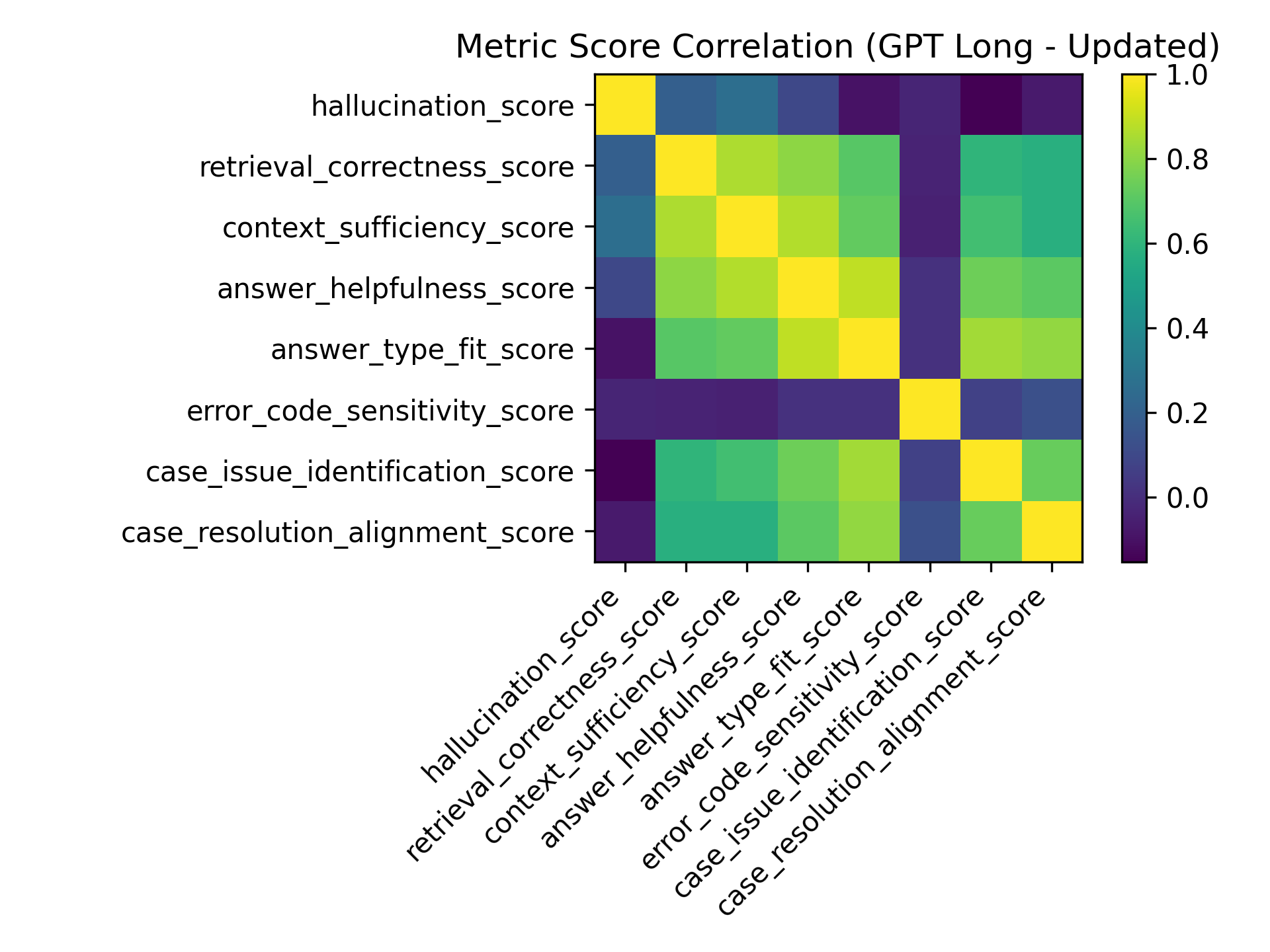}
    \caption{Correlation heatmap across evaluation metrics.}
    \label{fig:metric_correlation}
\end{figure}

\paragraph{Observations.}

Several important patterns emerge:

\begin{itemize}
    \item \textbf{Retrieval Correctness and Context Sufficiency} show moderate positive correlation, indicating that correct retrieval often—but not always—implies sufficient contextual grounding.
    \item \textbf{Case Issue Identification and Resolution Alignment} demonstrate strong correlation, suggesting that accurate issue diagnosis frequently leads to aligned resolution steps.
    \item \textbf{Hallucination} remains comparatively orthogonal to resolution alignment, reinforcing that factual grounding and actionable resolution are distinct evaluation dimensions.
\end{itemize}

This validates the need for multi-dimensional scoring rather than collapsing evaluation into a single faithfulness metric.

\subsection{Per-Metric Score Distributions}

We analyze score distributions across conversation turns.

\begin{figure}[t]
    \centering
    \includegraphics[width=\linewidth]{ 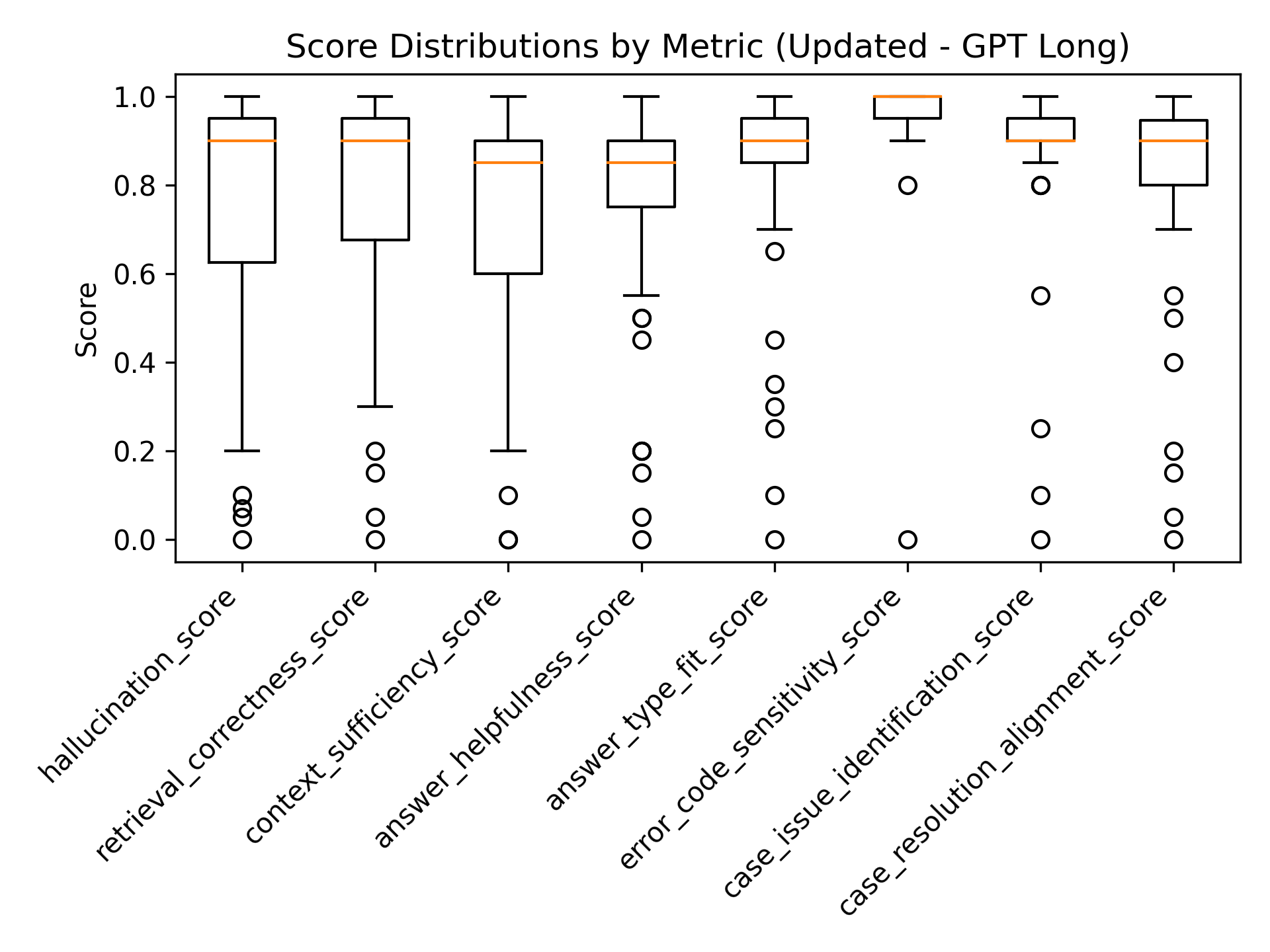}
    \caption{Score distributions per metric.}
    \label{fig:score_distributions}
\end{figure}

\paragraph{Distributional Insights.}

\begin{itemize}
    \item Retrieval-related metrics exhibit higher variance, particularly in early conversation turns involving ambiguous queries.
    \item Identifier integrity demonstrates bimodal behavior when structured identifiers (e.g., error codes, versions, commands) are present versus absent.
    \item Answer Type Fit maintains consistently high scores, suggesting structural compliance is easier to satisfy than deep contextual correctness.
\end{itemize}

These results illustrate that enterprise RAG systems fail more frequently in retrieval-context alignment than in structural formatting.

\subsection{Justification Length as a Complexity Proxy}

Because the framework requires textual justifications per metric, we analyze justification length as a proxy for reasoning complexity.

\begin{figure}[t]
    \centering
    \includegraphics[width=\linewidth]{ 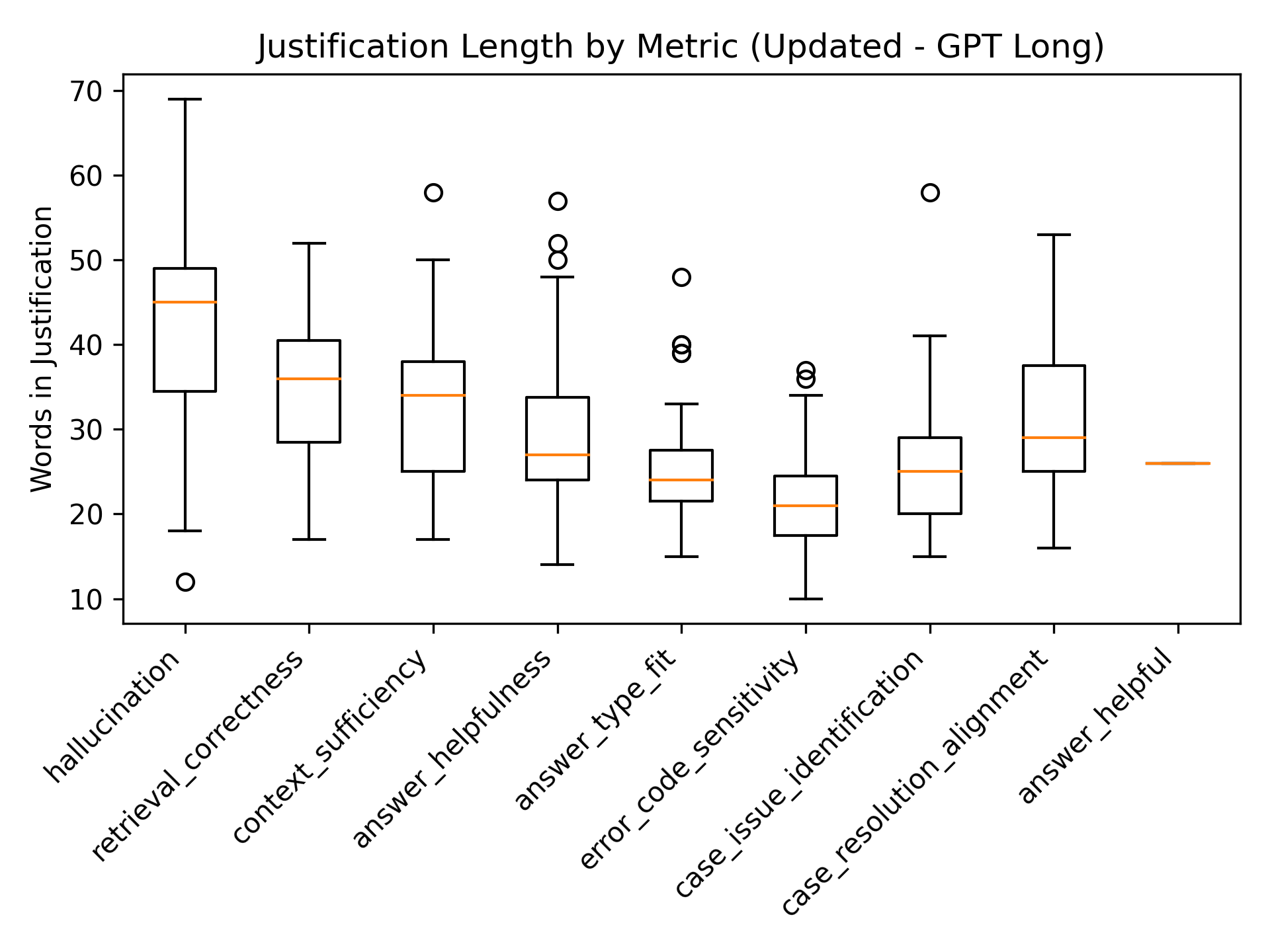}
    \caption{Word count distribution of LLM judge justifications per metric.}
    \label{fig:justification_lengths}
\end{figure}

\paragraph{Key Findings.}

\begin{itemize}
    \item Hallucination detection produces the longest justifications, consistent with the need for factual cross-verification.
    \item Answer helpfulness judgments require extended reasoning chains to evaluate completeness and actionability.
    \item Retrieval correctness explanations are comparatively shorter, reflecting binary factual matching.
\end{itemize}

This suggests that certain evaluation dimensions impose greater cognitive load on the judging model.

\subsection{Aggregate Metric Validity}

We examine the relationship between the weighted aggregate score ($S_{final}$) and answer helpfulness.

\paragraph{Construct Validity.}

Pearson correlation between $S_{final}$ and Answer Helpfulness was $r=0.82$ ($p<0.001$), indicating strong alignment
between the composite score and perceived answer utility. This indicates:

\begin{itemize}
    \item Improvements across structured enterprise metrics translate into improved perceived usefulness.
    \item The evaluation framework aligns with downstream user experience metrics.
    \item Composite scoring captures meaningful system quality signals.
\end{itemize}

This supports the framework’s construct validity in real-world enterprise settings.
\section{Extended Metric Interpretation}

\subsection{Retrieval and Sufficiency Distributions}

Mean retrieval correctness and context sufficiency scores were lower relative to identifier integrity and answer type fit. This reflects the strict rubric applied during evaluation.

Retrieval correctness required alignment with case-specific constraints such as product version, deployment context, and previously attempted steps. Partial matches or stale references were penalized. Context sufficiency required that retrieved evidence fully support safe resolution without omission of critical procedural elements.

In contrast, identifier integrity evaluates preservation and correct use of structured identifiers within the generated response itself (e.g., error codes, versions, commands), a narrower constraint that is typically easier to satisfy when explicit identifiers are present.

These distributional differences therefore reflect distinct failure modes rather than retriever collapse.
\subsection{Enterprise Metric Profile Visualization}
\begin{figure}[!htbp]
\centering
\includegraphics[width=0.9\columnwidth]{ 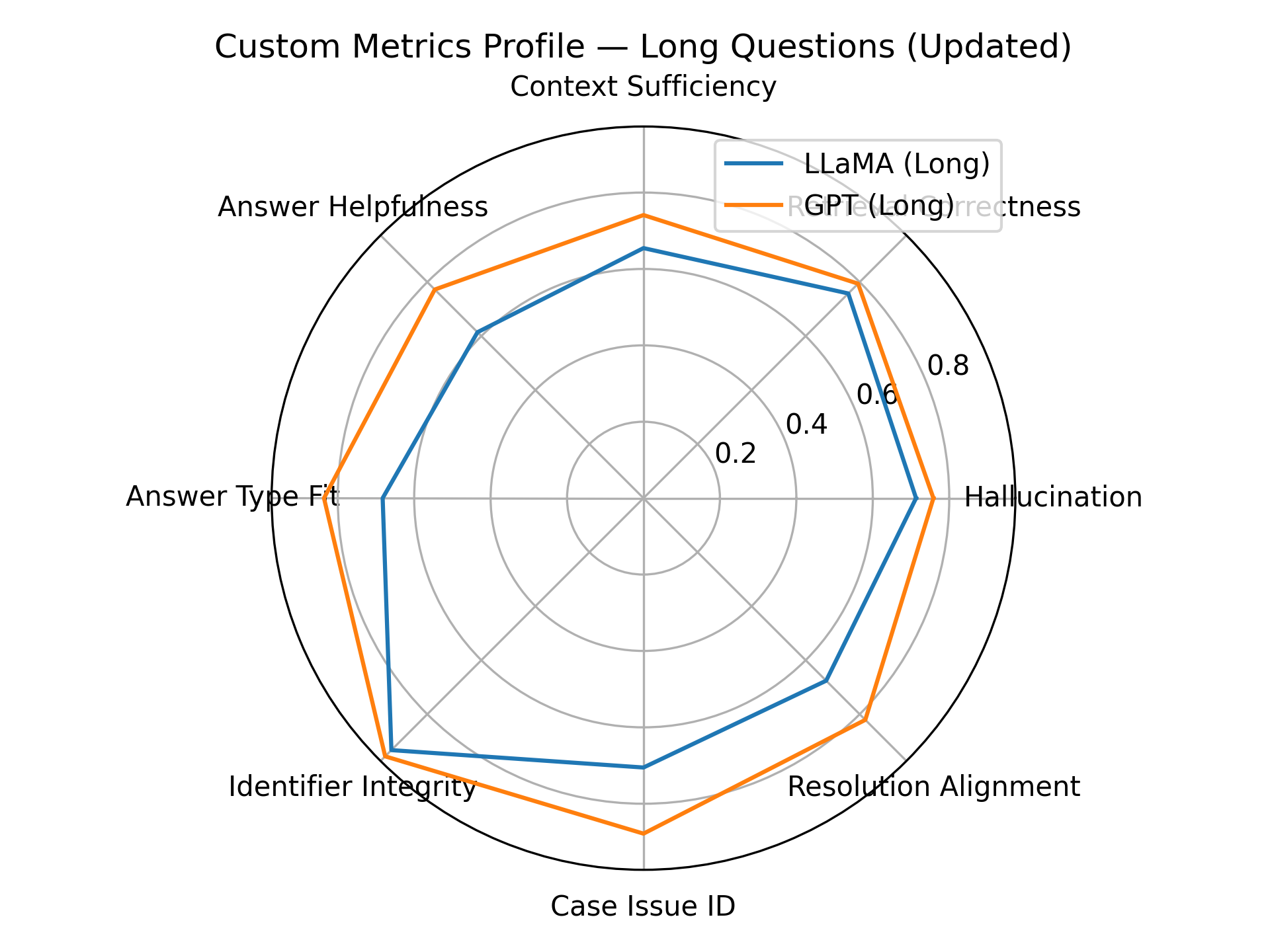}
\vspace{-0.4em}
\caption{
Radar visualization of normalized mean enterprise metric scores for long diagnostic queries.
While statistical significance is reported in Table~\ref{tab:significance}, 
this figure provides a qualitative view of performance divergence across workflow-aligned dimensions.
}
\label{fig:radar_appendix}
\vspace{-0.6em}
\end{figure}
\begin{figure}[t]
    \centering
    \includegraphics[width=\linewidth]{ 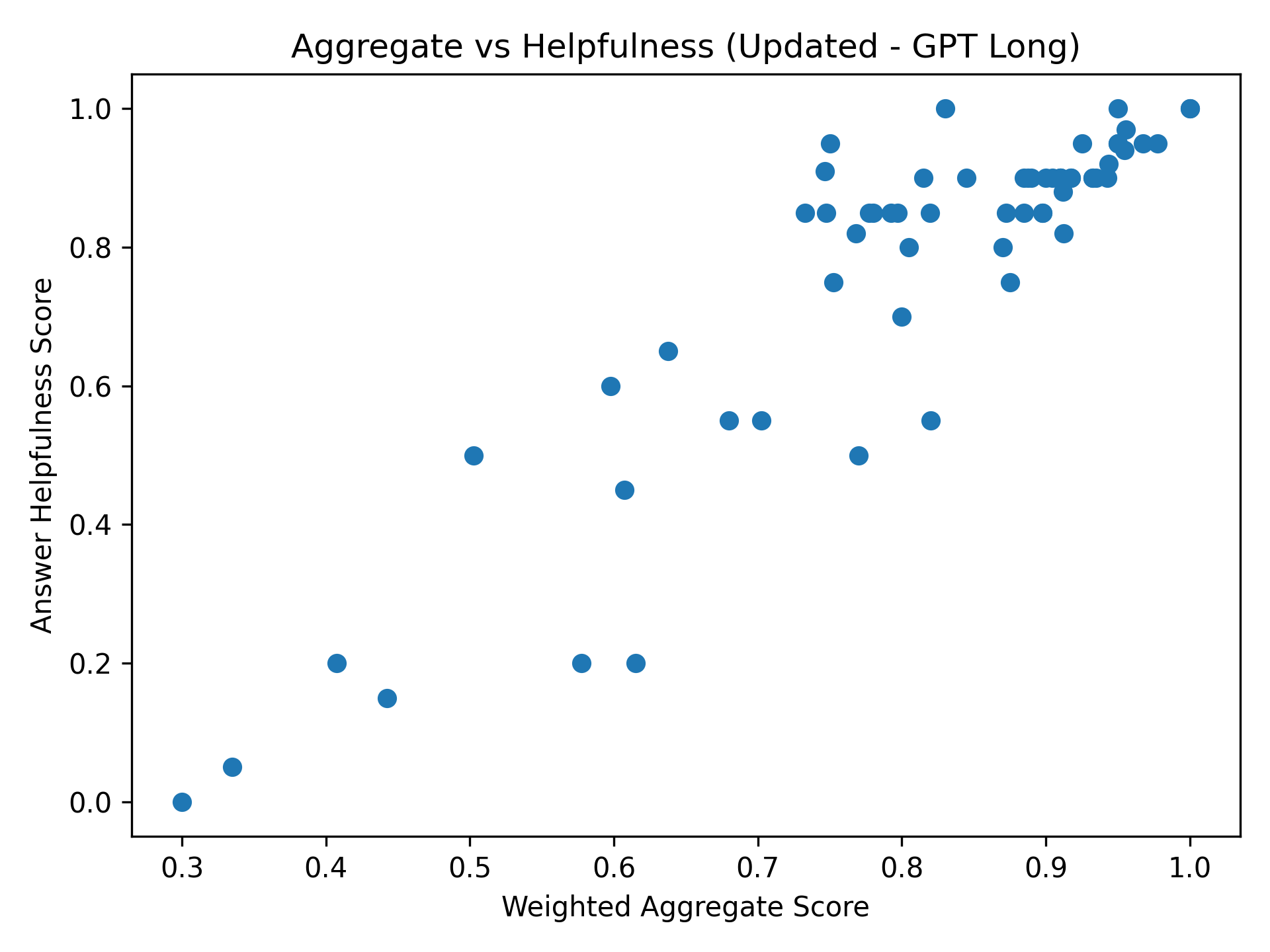}
    \caption{Weighted aggregate score ($S_{final}$) vs. answer helpfulness.}
    \label{fig:aggregate_vs_helpfulness}
\end{figure}

\section{Enterprise Motivation and Operational Context}
\label{sec:appendix_enterprise_motivation}

Enterprise RAG systems operate under operational constraints that differ substantially from benchmark-style QA evaluation. In customer support environments, partially helpful or structurally plausible responses can still lead to escalation, increased handling time, and reduced customer trust.

In production deployments, we observed cases where standard evaluation metrics rated responses as successful despite failures to:

\begin{itemize}
    \item Correctly interpret structured identifiers (e.g., error codes, versions)
    \item Follow prescribed troubleshooting order
    \item Align with established escalation workflows
    \item Maintain issue continuity across multi-turn exchanges
\end{itemize}

Such failures directly impact operational cost and deployment risk. Evaluation therefore functions not merely as an academic measurement tool, but as a gating mechanism for deployment, rollback decisions, and system iteration.

A key requirement in enterprise settings is \emph{diagnostic granularity}. When quality degrades, engineers must determine whether the root cause arises from:

\begin{itemize}
    \item Retrieval mismatch
    \item Insufficient contextual grounding
    \item Hallucination
    \item Case misinterpretation
    \item Resolution misalignment
\end{itemize}

Generic faithfulness and relevance metrics often conflate these dimensions, limiting their usefulness for production debugging. The proposed case-aware evaluation framework is motivated by repeated observations that precision-sensitive enterprise failure modes are underrepresented in standard RAG evaluation approaches.

\section{Extended Positioning Against Prior Work}
\label{sec:appendix_related}

\subsection{Limitations of Generic RAG Evaluation Assumptions}

Generic RAG evaluation often assumes: 
(i) context-independent queries,
(ii) complete and correct retrieved evidence, and 
(iii) externally verifiable ground truth.

Enterprise cases routinely violate these assumptions. Retrieval may be partially correct while omitting critical procedural steps. Evidence may mix stale and current policy documentation. Queries frequently depend on prior troubleshooting actions, environment constraints (e.g., production vs.\ lab), and organizational workflows.

In these settings:
\begin{itemize}
    \item Answer relevance may be high while case alignment is low.
    \item Faithfulness may be ambiguous when retrieved context is incomplete.
    \item Resolution quality depends on workflow compliance, not only factual correctness.
\end{itemize}

Our framework complements generic proxy metrics by explicitly modeling workflow compliance, precision integrity, and case resolution progress.

\subsection{Comparison to RAGAS, ARES, and RAGChecker}

RAGAS decomposes RAG quality into faithfulness and relevance signals, enabling reference-free evaluation \citep{es2023ragas,es2024ragasdemo}. ARES introduces a supervised pipeline for lightweight evaluators \citep{saadfalcon2024ares}. RAGChecker proposes a fine-grained diagnostic benchmark spanning retrieval and generation failures \citep{ru2024ragchecker}.

These works demonstrate the value of decomposed evaluation. However, enterprise deployments frequently require additional dimensions:

\begin{itemize}
    \item Structured identifier sensitivity (e.g., error codes)
    \item Case issue identification accuracy
    \item Compliance with prescribed troubleshooting order
    \item Multi-turn resolution alignment
\end{itemize}

These operational constraints motivate our case-aware metric design.

\subsection{Grounding and Attribution Literature}

Prior work studies factual grounding and attribution reliability. RARR introduces retrieval-driven revision pipelines \citep{gao2023rarr}. AttributionBench highlights evaluation challenges in attribution quality \citep{li2024attributionbench}. FActScore measures atomic factual precision in long-form text \citep{min2023factscore}.

While these approaches focus on factual support, enterprise RAG evaluation requires additional operational criteria—specifically, whether the answer advances resolution under case constraints.

\subsection{LLM-as-a-Judge Reliability Considerations}

LLM-as-a-judge approaches scale evaluation without exhaustive human labeling \citep{zheng2023mtbench}. However, studies document verbosity bias, position bias, and length sensitivity \citep{dubois2024lengthcontrolled}.

To mitigate such biases, our framework:
\begin{itemize}
    \item Enforces continuous scoring per metric
    \item Requires explicit justification
    \item Uses deterministic temperature settings
    \item Restricts inputs to retrieved evidence and case context only
\end{itemize}

This design improves auditability and reproducibility in enterprise evaluation settings.

\section{Evaluation Schema and Implementation Details}
\label{sec:appendix_schema}

\subsection{Evaluation Record Schema}

Each evaluated turn produces a structured JSON record containing:

\begin{itemize}
    \item real-valued scalar in $[0,1]$
    \item A short natural-language justification per metric
    \item An aggregated final score $S_{\text{final}}$
\end{itemize}

Table~\ref{tab:eval_schema_appendix} shows the full evaluation schema used in batch processing.

\begin{table*}[t]
\centering
\small
\begin{tabular}{lll}
\toprule
Metric & Score field & Justification field \\
\midrule
M1 Hallucination & \texttt{hallucination\_score} & \texttt{hallucination\_justification} \\
M2 Retrieval Correctness & \texttt{retrieval\_correctness\_score} & \texttt{retrieval\_correctness\_justification} \\
M3 Context Sufficiency & \texttt{context\_sufficiency\_score} & \texttt{context\_sufficiency\_justification} \\
M4 Answer Helpfulness & \texttt{answer\_helpfulness\_score} & \texttt{answer\_helpfulness\_justification} \\
M5 Answer Type Fit & \texttt{answer\_type\_fit\_score} & \texttt{answer\_type\_fit\_justification} \\
M6 Identifier Integrity & \texttt{identifier\_integrity\_score} & \texttt{identifier\_integrity\_justification} \\
M7 Case Issue Identification & \texttt{case\_issue\_identification\_score} & \texttt{case\_issue\_identification\_justification} \\
M8 Resolution Alignment & \texttt{case\_resolution\_alignment\_score} & \texttt{case\_resolution\_alignment\_justification} \\
\bottomrule
\end{tabular}
\caption{Full evaluation schema used in batch processing.}
\label{tab:eval_schema_appendix}
\end{table*}

\subsection{Continuous Scoring Interpretation}

Each metric is represented as a real-valued scalar in $[0,1]$. A score of 0 indicates complete failure for that dimension, while 1 indicates full compliance with the rubric. Intermediate values reflect partial satisfaction of the metric criteria.

Continuous scoring enables finer-grained aggregation, regression detection, and cross-case comparison in heterogeneous enterprise workflows. To improve stability, scoring anchors are defined in the rubric and deterministic prompting is used during evaluation.

\subsection{Batch Processing and Validation}

The evaluation pipeline enforces:
\begin{itemize}
    \item Deterministic prompting
    \item Strict JSON schema validation
    \item Automatic retries for malformed outputs
    \item Turn-level and case-level aggregation
\end{itemize}

\section{Detailed Limitations of Existing RAG Evaluation Frameworks}
\label{sec:appendix_limitations}

When applied to enterprise multi-turn deployments, existing RAG evaluation frameworks exhibit several recurring limitations:

\subsection{Independent Metric Evaluation}

Many frameworks score retrieval and generation quality independently. While useful for benchmarking, this separation can obscure compound failure modes in production systems, where partial retrieval mismatch combined with subtle reasoning drift produces operationally significant errors.

\subsection{Single-Turn Assumptions}

Enterprise cases frequently span multiple turns with clarification, hypothesis testing, and progressive resolution steps. Single-turn evaluation fails to capture degradation, recovery, and workflow continuity across turns.

\subsection{Lack of Severity Awareness}

Generic metrics often treat minor stylistic deviations and critical misinformation equivalently. In operational environments, severity matters: misinterpreting an error code or violating troubleshooting order carries substantially higher risk than minor phrasing issues.

\subsection{Limited Actionability}

Scores such as faithfulness or relevance provide coarse quality estimates but offer limited guidance for engineering decisions such as retriever adjustments, prompt revisions, or model rollouts. Enterprise teams require metrics that map directly to identifiable failure modes.

These limitations motivated the development of a case-aware, severity-sensitive evaluation framework tailored to operational RAG deployments.

\section{Extended Metric Definitions and Examples}
\label{sec:appendix_metric_details}

This section provides illustrative examples and edge-case clarifications for each metric.

\subsection{Hallucination vs. General Guidance}

General diagnostic advice is permissible if clearly framed as generic and not presented as case-specific fact. Hallucination occurs when the model invents case-specific assertions unsupported by retrieved evidence.

\subsection{Retrieval Failure Examples}

Common retrieval errors include incorrect product version, environment mismatch (production vs.\ lab), and stale policy references. These failures are isolated from generation quality by the retrieval correctness metric.
\subsection{Illustrative Failure Case: Workflow Violation Despite Correct Retrieval}

\textbf{Case Type:} Long diagnostic query involving a firmware update failure requiring prerequisite software upgrade.

\textbf{Retrieved Context (summarized):} The retrieved knowledge article specifies that software version 2.14 must be
installed before applying the firmware patch.

\textbf{Model Response (example failure):} The model recommends applying the firmware patch directly without verifying
or updating the software version.

\textbf{Expected Outcome:} The correct workflow requires verifying software version first and performing the prerequisite
upgrade before the firmware patch.

\textbf{Metric Impact:} Retrieval Correctness may remain high (context contains the correct prerequisite), while
Resolution Alignment is penalized due to violating documented sequencing. This illustrates why proxy metrics can
miss operational failure modes even when retrieval appears correct.

\subsection{Severity of Identifier Corruption}

Enterprise environments frequently depend on precise identifiers (error codes, command flags, file paths). Altered versions or corrupted commands constitute high-severity failures and are explicitly penalized.

\subsection{Workflow Misalignment}

Answers that repeat already attempted steps, skip required diagnostic sequencing, or recommend prohibited actions (e.g., unsafe production changes) reduce the Resolution Alignment score.

\section{Statistical Robustness Considerations}

\subsection{Variance Across Conversation Turns}

Evaluation variance is highest in:

\begin{itemize}
    \item Short queries lacking explicit error codes
    \item Multi-turn clarification exchanges
    \item Retrieval boundary cases with partially relevant documents
\end{itemize}

This highlights enterprise-specific RAG brittleness.
\subsection{Inter-Judge Consistency}
\label{app:judge_robustness}
We measure inter-judge rank consistency between GPT-4 and \texttt{llama-3.3-70b-instruct} on conversation-level scores using Spearman correlation.

\begin{table}[t]
\centering
\small
\begin{tabular}{lcc}
\toprule
Condition & $n$ & Spearman $\rho$ on $S_{\text{final}}$ \\
\midrule
Short (Llama responses) & 70 & 0.515 \\
Short (GPT-oss responses) & 70 & 0.577 \\
Long (Llama responses) & 63 & 0.691 \\
Long (GPT-oss responses) & 63 & 0.644 \\
\bottomrule
\end{tabular}
\caption{Inter-judge rank consistency between GPT-4 and Llama judges on weighted aggregate scores.}
\label{tab:interjudge_sfinal}
\end{table}

\section{LLM-as-a-Judge Prompt Design}

The evaluation prompt enforces strict schema constraints:

\begin{itemize}
    \item JSON-only output
    \item continuous score per metric
    \item Textual justification field per metric
    \item Explicit case-awareness conditioning
\end{itemize}

Case-awareness includes:

\begin{itemize}
    \item Prior turn context
    \item Error code presence
    \item Expected resolution trajectory
\end{itemize}

This ensures evaluation is conditioned on enterprise troubleshooting flows rather than isolated single-turn answers.

\section{Batch Evaluation Pipeline}

The evaluation pipeline includes:

\begin{itemize}
    \item Batched inference to reduce latency
    \item Automatic JSON schema validation
    \item Retry logic for malformed outputs
    \item Aggregation scripts for per-conversation scoring
\end{itemize}

All metrics are computed per turn and then aggregated to conversation-level summaries.

\section{Enterprise Failure Pattern Analysis}

The framework reveals failure modes not captured by generic RAG metrics:

\begin{itemize}
    \item Correct document retrieval but incorrect resolution sequencing
    \item Accurate issue identification without actionable steps
    \item Structurally valid responses lacking contextual grounding
    \item Over-generalized answers in error-code-driven scenarios
\end{itemize}

These patterns are critical in enterprise environments where resolution alignment directly impacts operational outcomes.

\section{Reproducibility and Deployment Considerations}

The framework is designed for deployment within enterprise governance constraints:

\begin{itemize}
    \item Deterministic evaluation configuration
    \item Auditable justification logs
    \item Batch-friendly architecture
    \item Model-agnostic scoring
\end{itemize}

All experiments were conducted with consistent prompt templates and temperature settings to ensure comparability across models.

\end{document}